\pdfoutput=1
\documentclass[10pt,final,conference,letterpaper]{ieeeconf}
\IEEEoverridecommandlockouts 
\usepackage{amsmath}

\usepackage{amsfonts}
\usepackage{pgf}
\usepackage{tikz}
\usepackage{bm}
\usepackage{enumerate}
\usepackage[T1]{fontenc}
\usepackage[noadjust]{cite}
\usepackage{subcaption}
\usepackage{algorithmic}
\usepackage{array,multirow,graphicx}
\usepackage[ruled, vlined, plain,linesnumbered]{algorithm2e}

\usepackage{fancyhdr}

\title{\LARGE \bf Planning on the fast lane: Learning to interact using attention mechanisms in path integral inverse reinforcement learning}

\author{Sascha Rosbach$^{1,2}$, Xing Li$^1$, Simon Gro{\ss}johann$^1$, Silviu Homoceanu$^1$ and Stefan Roth$^2$%
\thanks{$^{1}$The authors are with the Volkswagen AG, 38440 Wolfsburg, Germany
        {\tt\small \{sascha.rosbach, xing.li, simon.grossjohann, silviu.homoceanu\}@volkswagen.de}}%
\thanks{$^{2}$The authors are with the Visual Inference Lab,
        Department of Computer Science, Technische Universit\"at Darmstadt,
        64289 Darmstadt, Germany
        {\tt\small stefan.roth@visinf.tu-darmstadt.de}}%
}
\begin{document}
\maketitle
\thispagestyle{fancy} 

\begin{abstract}
General-purpose trajectory planning algorithms for automated driving utilize complex reward functions to perform a combined optimization of strategic, behavioral, and kinematic features.
The specification and tuning of a single reward function is a tedious task and does not generalize over a large set of traffic situations.
Deep learning approaches based on path integral inverse reinforcement learning have been successfully applied to predict local situation-dependent reward functions using features of a set of sampled driving policies.
Sample-based trajectory planning algorithms are able to approximate a spatio-temporal subspace of feasible driving policies that can be used to encode the context of a situation.
However, the interaction with dynamic objects requires an extended planning horizon, which depends on sequential context modeling.
In this work, we are concerned with the sequential reward prediction over an extended time horizon.
We present a neural network architecture that uses a policy attention mechanism to generate a low-dimensional context vector by concentrating on trajectories with a human-like driving style.
Apart from this, we propose a temporal attention mechanism to identify context switches and allow for stable adaptation of rewards.
We evaluate our results on complex simulated driving situations, including other moving vehicles.
Our evaluation shows that our policy attention mechanism learns to focus on collision-free policies in the configuration space.
Furthermore, the temporal attention mechanism learns persistent interaction with other vehicles over an extended planning horizon.
\end{abstract}

\section{Introduction}
To drive in complex environments, automated vehicles plan in spatio-temporal workspaces.
Sampling-based planning algorithms explore this workspace by sampling kinematically feasible actions.
Encoding features of dynamic objects is challenging because interaction occurs over an extended planning horizon.
Planning algorithms often rely on object predictions to derive features.
During persistent maneuvers such as lane changes, automated vehicles mediate between a set of rewards from kinematics, infrastructure, behavior, and mission.
A single reward function is often unable to evaluate a large set of heterogeneous driving situations.
In this work, we focus on situation-dependent reward predictions using inverse reinforcement learning (IRL) that enables persistent behavior over an extended time horizon.

However, two challenges arise regarding the spatial and temporal dimensions:
First, sampling a set of feasible driving policies often includes non-human-like trajectories that distort the assessment of the situational driving context.
Second, sequence-based reward prediction requires an efficient context encoding over an extended time horizon.
We propose a trajectory attention network that focuses on human-like trajectories to encode the driving context.
Furthermore, we use this context vector in a sequence model to predict a temporal reward function attention vector.
This temporal attention vector allows for stable reward transitions for upcoming planning cycles of a model-predictive control-based planner.

\begin{figure}  
  \vspace{1.8mm}
  \centering
  \includegraphics[width=0.48\textwidth]{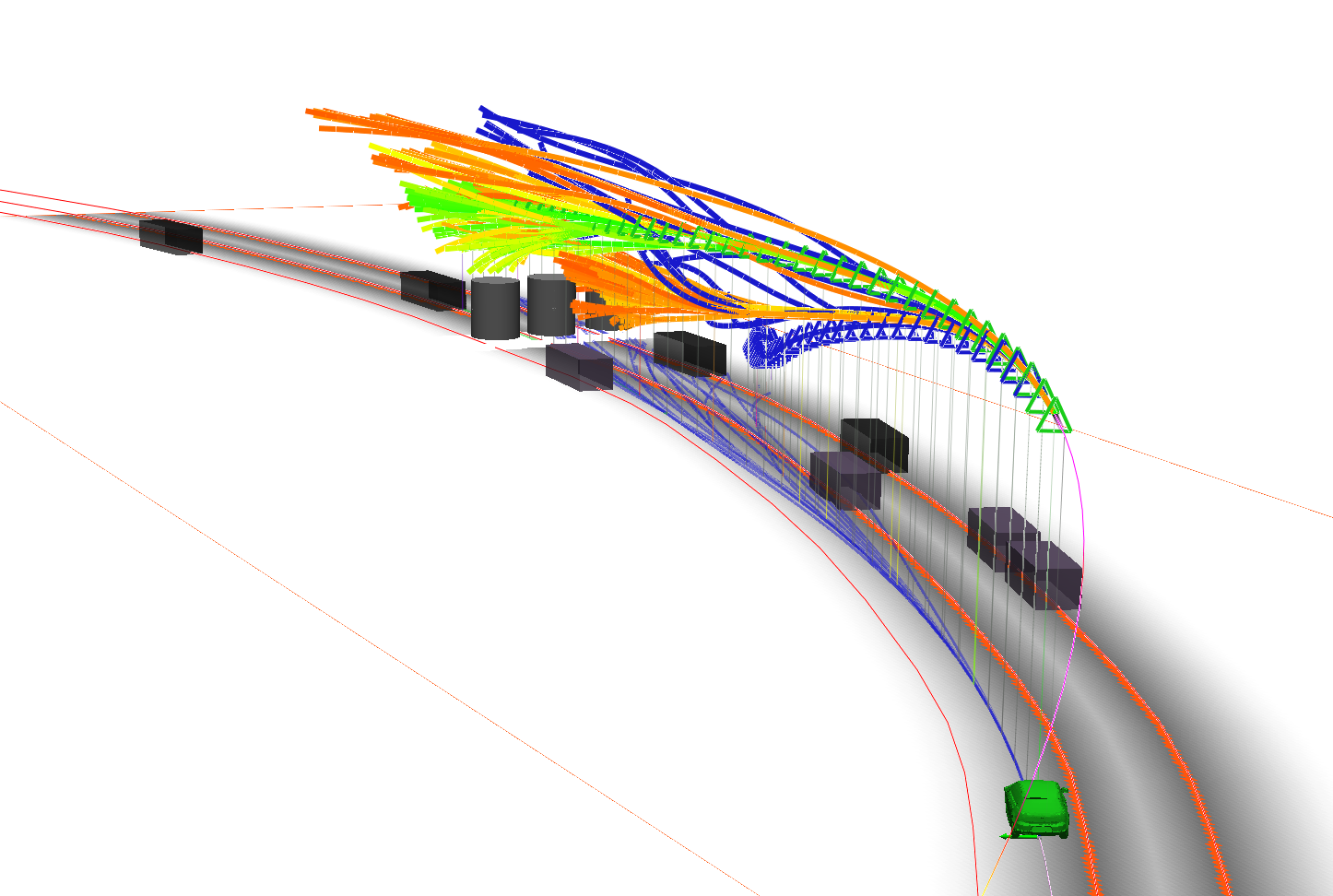}
  \caption{
    Illustration of our planner for automated driving, which samples policies for our deep inverse reinforcement learning approach.
    The z-axis corresponds to the velocity, whereas the ground plane depicts spatial feature maps such as distances from the lane centers.
    A subset of policies is visualized, where the green triangle shows the optimal policy, and the blue triangles high-light the highest policy attention.
    The color gradient corresponds to the policy value.
    Blue policies show high attention activation.
    The cylindric objects represent a stop barrier.
  }
  \label{fig:title}
\end{figure}

We evaluate the behavior of our approach in complex simulated driving situations over an oval course, including multiple lanes.
The agent has to reach checkpoints, stop at stop signs, and has to interact by passing other vehicles that drive at lower velocities.
We compare the reward predictions of our neural network architecture against baseline approaches using the expected value difference (EVD), expected distance (ED), and optimal policy distance (OPD) to the demonstrations.
Our experiments show that we are able to produce stationary reward functions if the driving task does not change while at the same time addressing situation-dependent task switches with rapid response by giving the highest weight to the reward prediction of the last planning cycle.

\section{Related Work}
General-purpose planning algorithms combine mission, behavior, and local motion planning.
These planning algorithms generate a set of driving policies in all traffic situations~\cite{mcnaughton2011phd}.
The policies are generated by sampling high-resolution actions based on action distributions that are derived from vehicle kinematics.
A sequence of sampled actions can produce driving policies with complex implicit maneuvers, e.g., double lane-changes and merges in the time gap between two vehicles.
The action sampling is achieved through massive parallelism on modern GPUs.
In contrast to classical hierarchical planning systems, these approaches do not decompose the decision-making based on behavior templates~\cite{heinrich2018phd}.
Thus, the planning paradigm does not suffer from uncertain behavior planning that is often introduced due to insufficient knowledge about the underlying motion constraints. 
However, general-purpose planning systems require a reward function that evaluates the policy set in terms of kinematic and environment features in all driving situations.
Specification and tuning of such a reward function is a tedious process that requires expert domain knowledge.
Motion planning experts often rely on linear reward functions, which do not generalize over a large set of driving situations.
The generalization of linear reward functions can be addressed by introducing an additional selection function of the final driving policy based on the generated policy set.
During the selection, clustering and reasoning techniques can be used to discover maneuver patterns and evaluate the final policy~\cite{gu2016iros}.
We adopt the methodology of a sample-based general-purpose planning algorithm and focus on predicting local situation-dependent reward functions to scale over a large set of driving situations.
In contrast to previous work, we do not use collision checking and features derived by post-sampling on the policy set~\cite{rosbach2019encoder}.
Instead, we challenge the deep learning approach to predict situation-dependent reward functions and thereby control the overall driving task.
Therefore, the interaction with infrastructure and dynamic vehicles is based on learned context representations.

In our previous work, we proposed a deep learning approach that predicts situation-dependent reward functions for such a sample-based planning algorithm.
These planning algorithms operate in a model-predictive framework to address updates of the environment~\cite{rosbach2019style, rosbach2019encoder}.
The deep learning approach based on IRL uses features and actions of sampled driving policies to predict a set of linear reward function weights.
The closed loop from sampled driving policies to reward function allows for dynamic updates of the reward weights over discrete planning cycles.
However, continuous reward function switches may result in non-stationary behavior over an extended planning horizon.
The authors found that the variance of the reward function prediction itself is proportional to the situational changes.
In this work, we concentrate on persistent interaction with other moving vehicles over an extended time horizon, which can only be achieved if temporally consistent reward functions are predicted.

Planning and reinforcement learning algorithms for automated driving often solve a Markov-Decision Process (MDP) to find an optimal action sequence.
The actions in automated driving are often represented as a tuple of wheel angle and acceleration.
Sutton et al. introduced a temporal abstraction to such primitive actions in semi-MDPs, which are referred to as options~\cite{sutton1999between}.
Options are closed-loop policies for taking actions over a period of time, e.g., stay on a lane, change a lane to the left or right~\cite{shalev-shwartz2016nips}.
Similar to the temporal driving abstraction in reinforcement learning that has been presented by Shalev et al.~\cite{shalev-shwartz2016nips}, we utilize temporal abstraction in IRL.
Previous work has investigated this hierarchical abstraction in IRL in terms of sub-task and sub-goal modeling using Mixture Models~\cite{krishnan2016hirl,vsovsic2018inverse}.
In contrast to this work, we utilize sequential deep learning models to determine task transitions automatically.

In order to interact with dynamic objects, the planning algorithm operates on a spatio-temporal space, where a subspace is sampled based on kinematic feasibility. 
Path integral features for a policy are approximated during the action-sampling procedure and describe features of individual policies.
In previous work, we focused on one dimensional (1D) convolutional neural network (CNN) architectures that generate a latent representation of trajectories~\cite{rosbach2019encoder}.
The situation-dependent context description is encoded in fully-connected layers using latent trajectory features of the 1D-CNN block.
The architecture's parameters largely depend on the policy set's size, which causes slow inference in recurrent models.
The size of the policy set used to understand the spatio-temporal scene can be significantly reduced by concentrating on relevant policies with a human-like driving style.
In this work, we use a policy attention mechanism to achieve this dimension reduction using a situational context vector.

Attention networks have gained significant interest in computer vision, natural language processing, and imitation learning~\cite{dzmitry2015translation,duan2017one, wang2019learning}.
Sharma et al. propose an attention-based model for action recognition in videos, which selectively focuses on parts of the video frames~\cite{sharma2015action}.
Fukui et al. use an attention branch to allow for visual explanation and improved performance in image recognition~\cite{fukui2019attention}.
We utilize the visual explanation capabilities of an attention mask to explain which of the sampled driving policies are most relevant in every planning cycle.
Wang et al. use an attention mechanism to learn unsupervised object segmentation~\cite{wang2019learning}.
They leverage the availability of affordable eye-tracking from human gazes to annotate objects.
Similar to this work, we use odometry records as affordable labels to add supervised conditions on our situational context vector.
Thereby, high attention on trajectories yields a proxy for closeness to expert demonstrations.

\section{Preliminaries}

Planning is often formulated as an MDP consisting of a 5-tuple ${\mathcal{S},\mathcal{A},T,R,\gamma}$, where $\mathcal{S}$ denotes the set of states, and $\mathcal{A}$ describes the set of actions.
In the domain of continuous control, an action $a$ is integrated over time $t$ using a transition function $T(s,a,s')$ for ${s, s' \in \mathcal{S}, a \in \mathcal{A}}$.
Every action $a$ in state $s$ is evaluated using a reward function $R$ that is discounted by $\gamma$ over time $t$.
The reward function uses features that are computed using an environment model and a vehicle transition model.
The planner explores the subspace of feasible policies $\Pi$ by sampling actions from a distribution conditioned on vehicle dynamics for each state $s$.
The reward function is a linear combination of $k$ static and kinematic features $f_i$ with weight $\theta_i$ such that  ${\forall(s,a) \in \mathcal{S} \times \mathcal{A}: R(s,a)=\sum_{i \in K} {-\theta_i f_i{(s,a)}}}$.
The value of a policy $V^{\pi}$ is the integral of discounted rewards during continuous transitions.
The feature path integral $f^{\pi}_i$ for a policy $\pi$ is defined by ${f_i^{\pi}=\int_t\gamma_t f_i(s_t,a_t)\,dt}$.
We project odometry records $\zeta$ of expert demonstrations into the state-action space to formulate a demonstration policy based on a Euclidean distance metric ensuring $\pi^D \in \Pi$.
To extend the temporal planning horizon, a sequence of a-priori unknown reward functions can be defined as $R_{seq} = [R_{seq}^{(1)},...,R_{seq}^{(k)}]$.
Similar to options in a semi-MDP, which are a generalization of primitive actions, a task can be decomposed into a sequence of subtasks, which depends on a preceding sequence $R_{seq}^{(k-1)}$.
Thereby planning can be described in an MDP within a set of MDPs \mbox{$\mathcal{M} = [M^{(1)},...,M^{(k)}]$}, each having different reward functions $R^{(k)}$.

\subsection{Maximum entropy PI deep IRL}

IRL allows finding the reward function weights $\bm{\theta}$ that enable the optimal policy $\pi^*$ to be at least as good as the demonstrated policy $\pi^D$~\cite{arora2018}.
The behavior of a demonstration is thereby indirectly imitated by the planning algorithm~\cite{ng2000icml}.
In path integral (PI) IRL, we formulate a probabilistic model that yields a probability distribution over policies, $p(\pi|\bm{\theta})$~\cite{aghasadeghi2011iros,theodorou2010generalized}.
For each planning cycle, we optimize under the constraint of matching the expected PI feature values $\mathbb{E}_{p(\pi|\bm{\theta})}[\bm{f}^{\pi}]$ of the policy set $\Pi$ and the empirical feature values $\hat{\bm{f}}^{\Pi^D}$ of the demonstrations.
Imperfect demonstrations introduce ambiguities in the optimization problem, which Ziebart et al.~\cite{ziebart2008aaai} propose to solve by maximizing the entropy of the distribution.
The policy distribution is given by
\begin{align}
  \label{eq:policyprob}
  p(\pi|\bm{\theta})= \dfrac{1}{Z}\exp(-\bm{\theta}^\top\bm{f}^{\pi}).
\end{align}

Due to the exponential growth of the state-action space it is often intractable to compute the partition function \mbox{$Z=\sum_{\pi \in \Pi}{\exp(-\bm{\theta}^\top \bm{f}^{\pi})}$}.
We approximate the partition function by sampling driving policies similar to Markov chain Monte Carlo methods.
Maximizing the entropy of the distribution over policies subject to the feature constraints from demonstrated policies implies that the log-likelihood $L(\bm{\theta})$ of the observed policies under the maximum entropy distribution is maximized.
In previous work, we formulated a deep learning approach for PI maximum entropy IRL, which approximates a complex mapping between PI features $\bm{f}^{\Pi}_k$, actions $\bm{a}^{\Pi}_k$  and reward function weights $\bm{\theta}_{k+1}$ at MPC cycles $k$, given by $\bm{\theta}_{k+1} \approx g(\bm{\Theta},\bm{f}_k,\bm{a}_k)$.

The IRL problem can be formulated in the context of Bayesian inference as maximum a posteriori estimation, which entails maximizing the joint posterior distribution of observing expert demonstrations $\Pi^D$.
We calculate the maximum entropy probability based on the linear reward weights $\bm{\theta}$, which are inferred by the network with parameters $\bm{\Theta}$ as

\begin{equation}
  L(\bm{\theta}) = L(g(\bm{\Theta},\bm{f},\bm{a})) =\sum_{\pi^D \in \Pi^D}\ln{p(\pi^D|g(\bm{\Theta},\bm{f},\bm{a}))}.
  \label{MLE}
\end{equation}

The gradient for the log-likelihood $L(\bm{\theta})$ can be calculated in terms of $\bm{\Theta}$ as
\begin{equation}
  \begin{split}
    \frac{\partial L}{\partial \bm{\Theta}} & =\frac{\partial L}{\partial \bm{\theta}} \cdot \frac{\partial \bm{\theta}}{\partial \bm{\Theta}}\\
    & = \Big[\sum_{\pi \in \Pi} p(\pi|\bm{\theta})\bm{f}^\pi - \bm{\hat{f}}^{\Pi^D}\Big] \cdot\frac{\partial}{\partial \Theta}{g(\bm{\Theta},\bm{f},\bm{a} )}.
  \end{split}
  \label{eq:deepirlgrad}
\end{equation}

The gradient is separated into the maximum entropy gradient in terms of $\bm{\theta}$ and the gradient of $\bm{\theta}$ w.r.t.~the network parameters $\bm{\Theta}$, which can be directly obtained via backpropagation in the deep neural network.

\subsection{Open-loop reward learning}
Training IRL algorithms is often time-consuming.
The MDP has to be solved with respect to the current reward function in the inner loop of reward learning.
We reduce the time constraint by running our planning algorithm prior to training with a randomly initialized reward function $\bm{\theta}_0$.
This allows us to generate a buffer of policy sets with corresponding features and actions.
Sampling high-resolution actions allows us to project odometry records $\zeta$ in the state-action space.
We use a weighted Euclidean distance metric calculation in the sampling procedure to evaluate distances of policies to the odometry of the expert trajectories~\cite{rosbach2019style}.
The weighted Euclidean metric was evaluated against sequence alignment methods such as dynamic time warping and found to be sufficient to select demonstration policies.
In addition to the path integral features $\bm{f}^{\pi}$ of a policy, we use features of the policies at time-equidistant control points $\bm{c}^{\pi}$.
The feature sequence includes the lateral offsets with reference to the ego position, yaw, and progress along the route.
The progress is calculated using Dijkstra's algorithm on the road-network after receiving the target destination.
The training algorithm is run for a predefined number of epochs, ensuring that the convergence metrics, the EVD and ED, reach the desired threshold value.
For each epoch, the training dataset is shuffled and divided into batches to perform mini-batch gradient descent.

\begin{figure*}[ht]
  \vspace{4mm}
    \includegraphics[width=\textwidth]{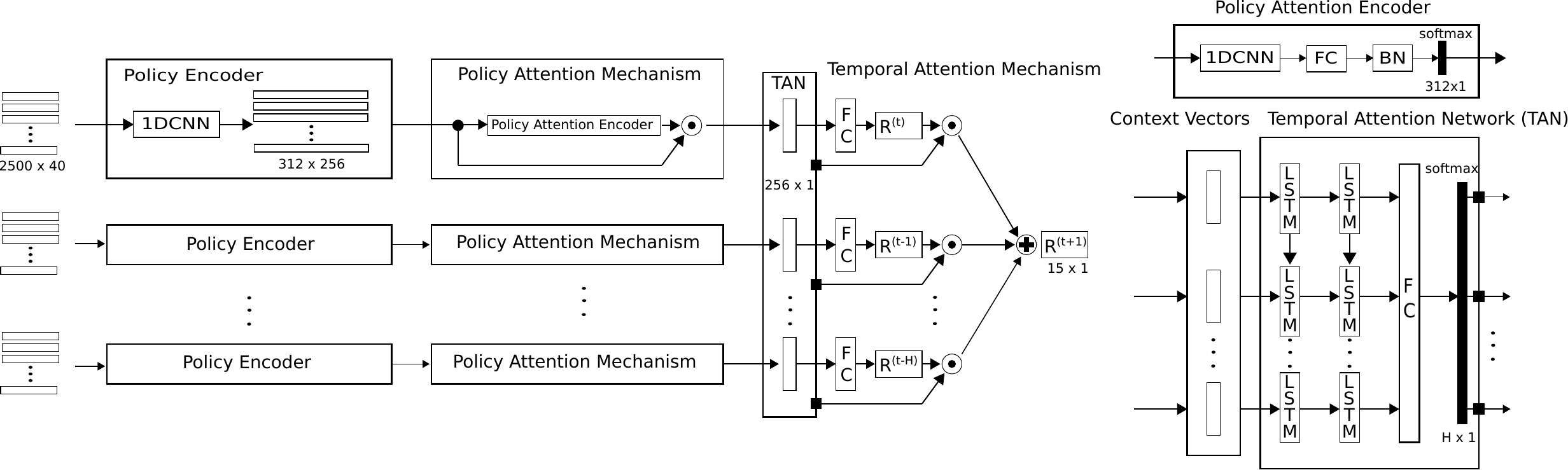}
  \caption{
    Neural network architectures for situation-dependent reward prediction.
    Policy temporal attention architecture consisting of policy attention and temporal attention mechanism.
    Inputs are a set of planning cycles, each having a set of policies.
    Policy encoder generates a latent representation of individual policies.
    Policy attention mechanism produces a low-dimensional context vector, which is forwarded to the temporal attention network (TAN).
    Policy temporal attention mechanism predicts a mixture reward function given a history of context vectors.
  }
    \label{fig:nn}
\end{figure*}

\section{Neural Network Architecture}
We propose a deep learning architecture for PI deep IRL.
This architecture uses inputs of PI features $f^\Pi$, actions $a^\Pi$, and a sequence of spatio-temporal features in the form of policy control points $c^\Pi$.
Our deep IRL architecture is separated into a policy attention mechanism and a temporal attention mechanism, as shown in Fig.~\ref{fig:nn}. 

\subsection{Policy Attention}
The policy attention mechanism generates a 1D context vector of the situation.
We feed the policy sets into a policy encoder, which relies on 1DCNN layers to generate latent features of individual policies.
The combined policy encoder and policy attention mechanism are referred to as policy attention CNN (PACNN).
A policy attention encoder uses combinations of 1D convolutions, average pooling, and fully-connected layers to compute a policy attention vector.
Our attention vector is based on the soft attention mechanism~\cite{dzmitry2015translation}.
We perform a softmax operation over the output of the attention encoder network to generate a 1D vector.
The attention vector essentially filters non-human-like trajectories from the policy encoder.
We combine the maximum entropy IRL gradient with a semi-supervised attention loss~\cite{wang2019learning}.
The semi-supervised loss is based on the mean absolute distance towards the expert demonstration.
To compute the loss, we sort the policies in ascending order of progress along the route.
Sorting enables a consistent relationship between attention loss and the sampled policy set distribution.
The output of spatial attention is multiplied by a learned scalar~\cite{zhang2018self}.
The scalar learns cues in the local neighborhood and gradually assigns more weight to non-local evidence.
The maximum entropy gradient is calculated based on the policy set of the input distribution~\cite{rosbach2019encoder}.
We use 1D average upsampling of the attention vector to match the dimensionality of policy sets.
This allows us to visualize the trajectory attention during inference.

\subsection{Temporal attention}

In a second training step, we use context vectors of our PACNN networks and the corresponding situation-dependent reward functions to predict the reward functions for the next planning cycle at time $t+1$.
We do so by taking a sequential history size $H$ of context vectors and reward functions into account.
The temporal attention network consists of a two-layered recurrent long short-term memory (LSTM) network and a fully-connected network of four layers.
The output is a 1D weight vector computed by a softmax activation function.
The final reward function is a mixture of situation-dependent reward functions $R^{(t+1)} = \sum_{h \in H} w_{h} R^{(h)}$.
In contrast to the PACNN network, the temporal attention network PTACNN learns to predict by training with the maximum entropy gradient of the future planning cycle at time $t+1$.
This architecture allows for long sequence lengths and fast inference during the prediction due to a low dimensional context vector.
The overall idea is similar to expectation-maximization (EM) IRL, which uses a mixture of clustered reward functions to infer a situation-dependent reward function given features of the demonstrations~\cite{babesApprenticeshipLearningMultiple2011}.
In contrast to the mixture model, we infer a mixture of sequential reward functions based on a latent context description of the situations.

\section{Experiments}
We conducted our experiments on complex simulated scenarios.
The driving situations on our oval course are designed in a way that a lap completion requires continuous task predictions.
The oval map includes multiple lanes, as depicted in Fig.~\ref{fig:title}.
Four checkpoints provide a proxy for the target locations on the course; checkpoints are toggled from inner to outer lanes to enforce mission-oriented lane-changes.
There are multiple exits on the oval, which make the mission evaluation a requirement.
On two locations of the oval, stop signs span over all lanes to assess stopping, starting, and making progress along the route.
At most 15 vehicles are spawned at random at a distance of 200~m from the ego vehicle.
The vehicles drive with constant velocity if they do not interact with other vehicles or infrastructure.
The spawning velocity is selected at random in the range of 25~-~35~kph.
The agent's target velocity is set to 70~kph, which requires constant mediation between strategic, behavioral, and motion-related reward features.
Due to the large velocity difference between the agent and other vehicles, it is expected that the agent drives aggressively.
In the presence of 15 other vehicles, this implies that the agent has to learn how to merge into small gaps between vehicles and pass without colliding.

\subsection{Data collection and simulation}

We collect expert driving demonstrations by recording the optimal policies of an expert-tuned planning algorithm.
The expert-tuned planner uses a manually tuned reward function and a model-based trajectory selection.
Similar to Gu et al.~\cite{gu2016iros}, the expert-tuned planner uses topological clustering and additional features that are computed on the policy set to derive the final driving policy.
A crucial input for the selection is the progress value of policies along the route.
This feature gets the vehicle moving and influences mission-oriented lane-changes.
Once the odometry of the expert-tuned planning algorithm is recorded, the model-based selection, and its additional features are disabled.
This is done to test if learned context vectors are able to encode latent features of the policy set, which allow the indirect expert-planner imitation.
During data collection, the odometry of the expert-tuned optimal policies are recorded.
We utilize the same data collection principle, as in \cite{rosbach2019style,rosbach2019encoder}.
The odometry records are projected into the state space to formulate geometrically close demonstrations $\pi^D$ for the training, validation, and test dataset.
For our training datasets, we do not assume prior knowledge of the reward function, therefore solve the MDP using a random reward function.
For our tests on sequential datasets, we collect policy sets using the expert-tuned planner.
By projecting the odometry of the expert optimal policies in state space during testing, we achieve a proxy for perfect imitation.

\subsection{Reward feature representation}
The reward function features are computed during the action sampling procedure and describe vehicle motion, infrastructure, and time-dependent distances to objects.
We consider 15 manually engineered features.
Infrastructural features are derived from street networks~\cite{homeier2011itsc}.
Derivatives of lateral and longitudinal actions describe the vehicle kinematics.
Lane change dynamics are described by lane change delay and lateral overshooting.
The lane change delay punishes performing lane changes at the end of the planning horizon.
Spatio-temporal proximity is calculated from object motion predictions.

\subsection{Baseline approaches}

All of the baseline approaches use the path integral IRL training methodology and produce a linear combination of reward weights.
We consider a linear IRL (LIRL) approach and two non-recurrent deep IRL neural network architectures as baseline methods.
The neural networks generate a latent context representation of the input policy distribution.
The 1DCNN architecture uses fully-connected layers to encode the context from latent policy features; we refer to this architecture as 1DCNN~\cite{rosbach2019encoder}.
An alternative architecture uses 1D convolutions over latent features to decrease the neural network parameters.
This architecture is referred to as Bi1DCNN. 

\begin{figure*}
  \centering
  \vspace{2mm}
            \begin{subfigure}{0.32\textwidth}
              \centering
              \includegraphics[width=\textwidth]{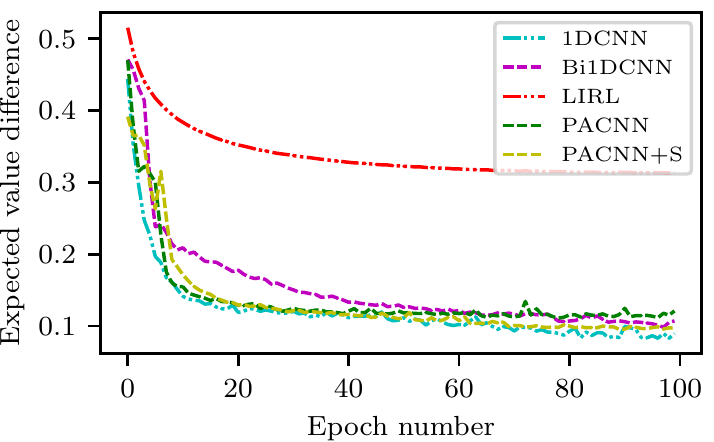}
              \caption{
                Training: Convergence on a non-sequential training dataset based on EVD.
              }
              \label{fig:evd}
            \end{subfigure}
            ~
            \begin{subfigure}{0.32\textwidth}
              \centering
              \includegraphics[width=\textwidth]{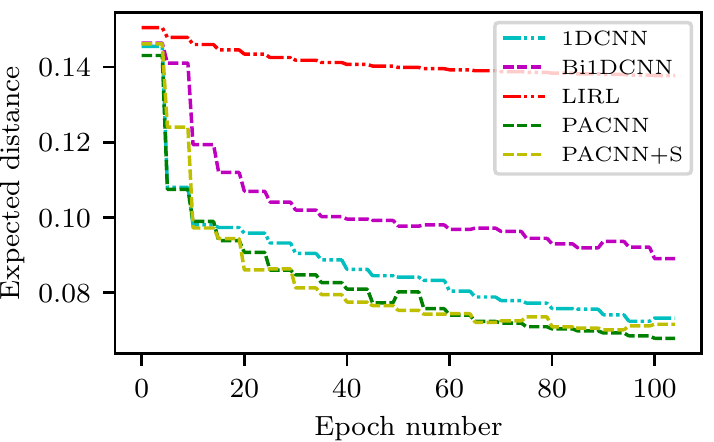}
              \caption{
                Validation: Convergence on a sequential validation dataset based on ED.
              }
              \label{fig:ed}
            \end{subfigure}
            ~
            \begin{subfigure}{0.32\textwidth}
              \centering
              \includegraphics[width=\textwidth]{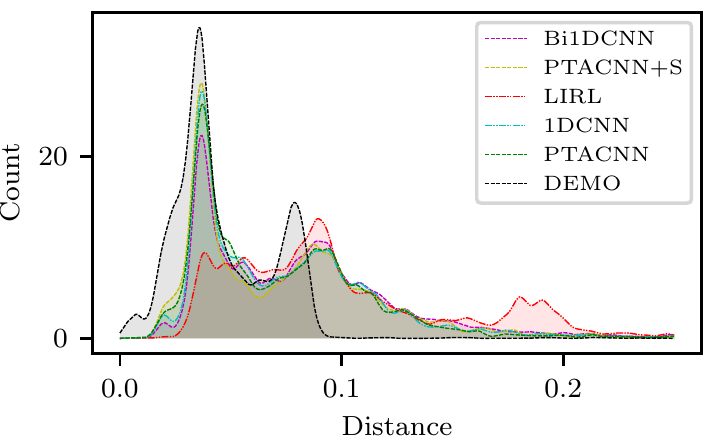}
              \caption{
                Test: Distance of policy to the demonstration on a sequential test dataset.
              }
              \label{fig:dist}
            \end{subfigure}
            \caption{
              Training and test results of our proposed methods in contrast to baseline approaches.
              \emph{(a)} Convergence on a non-sequential training dataset based on EVD.
              \emph{(b)} Convergence on a non-sequential validation dataset based on ED.
              \emph{(c)} Distance distribution of optimal policies to the expert demonstrations on a sequential test dataset.
              The lower bound for the distance distribution is given the distance of the demonstration (DEMO).
              During the sequential prediction all deep learning approaches use a history size $H=10$.
              }
            \label{fig:trainvaltest}
\end{figure*}

\begin{table}
  \vspace{1.8mm}
  \centering
\caption{
Overview of average test performance based on ED, and OPD.
Tests are conducted on a test dataset, recorded by an expert-tuned planning algorithm.
Number of trainable variables of the neural networks are listed and split for the PTACNN networks into the policy (left) and temporal (right) attention network parameters.
}
\begin{tabular}{l|c|c|c}
  Approach           &   Trainable variables                     & $\overline{ED}$      & $\overline{OPD}$ \\ \hline
  LIRL               &   15                             & 0.121                & 0.116 \\
  Bi1DCNN            &   18.7~M                         & 0.105                & 0.096\\                                                                          
  1DCNN              &   22.6~M                         & 0.094                & 0.088\\                                                                          
  \textbf{PTACNN}    &   2.94~M + 1.61~M = 4.55~M       & \textbf{0.092}       & \textbf{0.086}\\                                                                          
  \textbf{PTACNN+S}  &   2.94~M + 1.61~M = 4.55~M       & \textbf{0.091}       & \textbf{0.081}\\                                                                          
\end{tabular}
\label{fig:test}
\end{table}

\section{Evaluation}

We evaluate the performance of our proposed spatio-temporal attention networks against our baseline approaches.
First, we evaluate the context encoding capabilities of the approaches.
Here, we concentrate on the convergence of our PACNN network against neural-networks without such an attention mechanism.
The convergence is analyzed in terms of EVD and ED on training and validation datasets over training epochs.
Second, we compare the sequential prediction performance in terms of the optimal policy distance (OPD) to the expert-demonstration on a playback test dataset.
Furthermore, our supplementary video displays the closed-loop reward function prediction and driving performance in challenging driving situations.

\subsection{Comparison with expert-demonstrations}

Fig.~\ref{fig:trainvaltest} depicts the training, validation, and test results of our evaluated methods.
All methods in the convergence plot are trained using the maximum entropy gradient of the trajectory input distribution.
During validation and testing, we calculate the EVD, ED, OPD based on the inferred reward function using a history size $H=10$ for all methods.
This means that all methods except the PTACNN use a mean of inferred reward weights over the history size.
We configured the planning algorithm so that it yields approximately 2.500 policies during each planning cycle.
Fig.~\ref{fig:evd} represents the convergence of our training, which is measured by EVD over epochs~\cite{rosbach2019style}.
In the EVD calculation, the value is normalized by the value of the demonstration, since the weights may increase their range over the training epochs.
We abort training after achieving a high ED and EVD reduction and observe the weight distributions over the epochs.
In our training dataset, we use one hour of driving demonstrations, which provide approximately 17.000 planning cycles and an equal amount of expert-demonstrations $\pi^D$.
We split our evaluation dataset into validation and hold out test dataset.
PACNN+S and PTACNN+S have been trained using an additional semi-supervised loss based on the mean absolute policy distance to the demonstration.
We calculate the EVD every epoch and perform validation every fifth epoch.

All deep IRL methods converge to a similar EVD, in contrast to LIRL, which is unable to fit a single reward function yielding low EVD.
Bi1DCNN converges after 100 epochs of training with an ED of 0.1.
PACNN, PACNN+S, and 1DCNN converge at a close ED proximity at a value of 0.07.
Using a semi-supervised loss in addition to the maximum entropy gradient did not improve nor decrease the training results in terms of EVD significantly.
All deep IRL approaches show a similar peak in the OPD distribution, akin to the demonstration, as depicted in Fig.~\ref{fig:dist}.
The distance distribution of the demonstration is the lower bound for the OPD.
The nonzero lower bound is caused by discretization errors of the state-action space and the planner's nondeterministic optimization methodology and can be regarded as a proxy for perfect imitation.
In addition to the distribution, we summarize the test results in Table~\ref{fig:test}.
PTACNN+S is trained in a second stage using the context vector and reward predictions of PACNN+S.
The generalization of a single reward function is not achieved, as shown in the ED reduction and OPD on the test set.
The performance of 1DCNN and Bi1DCNN models on the validation set is proportional to the trainable variables after latent feature extraction using 1DCNNs.
1DCNN uses fully-connected layers to learn a context representation.
In contrast to PACNN, Bi1DCNN learns a set of filters over latent variables of policies.
The attention networks stand out, having fewer parameters and a low-dimensional context vector while yielding similar performance as compared to larger neural network architectures.
PACNN uses eight times less trainable variables than 1DCNN and six times less trainable variables than Bi1DCNN.
PTACNN performs best on the test dataset, yet the evaluation of persistent reward predictions using temporal attention requires a closed-loop inference.

\subsection{Visualization of attention mechanisms}
During inference, the attention mechanisms can be visualized, as depicted in Fig.~\ref{fig:title}. 
The blue trajectories show the trajectories with the highest policy attention activation.
A color gradient is assigned to the policy value that ranges from green (high) to red (low).
Only a subset of all feasible policies is visualized.
Our video shows the driving performance during closed-loop inference of our proposed method.
In the video, we display two additional figures.
A radar chart depicts the predicted reward function weights, and a bar chart shows the temporal attention activation. 
PTACNN is able to control the complete driving task and interacts with other vehicles without relying on model-based collision checking.
  
\section{Conclusion and Future Work}
In this work, we propose a deep network architecture that is able to predict situation-dependent reward functions for a sample-based planning algorithm.
Our architecture uses a temporal attention mechanism to predict reward functions over an extended planning horizon.
This is achieved by generating a low dimensional context vector of the driving situation from features and actions of sampled-driving policies.
Our experiments show that our attention mechanisms outperform our baseline deep learning approaches during comparisons against expert-demonstrations. 
In closed-loop inference, our approach is able to control the complete driving task in challenging situations while only learning from one hour of driving demonstrations.
In future, we plan to train the algorithm on a large scale dataset and use it in combination with model-based constraints in real-world driving situations.
Furthermore, we want to integrate raw sensory data into the deep inverse reinforcement learning approach so as to learn relevant features of the environment automatically.

\newpage
\bibliographystyle{IEEEtran}
\bibliography{bib/conf_names_abrv,bib/library}
\end{document}